\documentclass[sigconf]{acmart}
\AtBeginDocument{%
  \providecommand\BibTeX{{%
    \normalfont B\kern-0.5em{\scshape i\kern-0.25em b}\kern-0.8em\TeX}}}

\setcopyright{acmcopyright}
\copyrightyear{2023}
\acmYear{2023}
\acmDOI{XXXXX}

\acmConference[AdKDD 2023]{}{Long Beach}{CA}
\acmPrice{}
\acmISBN{}

\usepackage{times}
\usepackage{epsfig}
\usepackage{graphicx}
\usepackage{amsmath}
\usepackage{subcaption}
\usepackage{comment}

\begin{document}


\title{Staging E-Commerce Products for Online Advertising \\ using Retrieval Assisted Image Generation}

\begin{comment}
\author{
Yueh-Ning Ku \\
Yahoo Research\\
Sunnyvale\\
{\tt\small yuehning.ku@verizonmedia.com}
\and
Mikhail Kuznetsov \\
Yahoo Research\\
NYC \\
{\tt\small kuznetsov@verizonmedia.com}
\and
Shaunak Mishra \\
Yahoo Research\\
NYC \\
{\tt\small shaunakm@verizonmedia.com}
\and
 Paloma de Juan\\
Yahoo Research\\
NYC \\
{\tt\small pdjuan@verizonmedia.com}
}

\end{comment}


\author{Yueh-Ning Ku}
\authornote{Work done while at Yahoo Research.}
\authornotemark[1]
\affiliation{%
%  \institution{}
  \country{Snap Inc., USA}
 }

\author{Mikhail Kuznetsov}
\authornote{Work done while at Yahoo Research.}
\affiliation{
  \country{Amazon, USA}
%  \institution{Amazon}
}
\author{Shaunak Mishra}
\authornote{Work done while at Yahoo Research.}
\affiliation{
  \country{Amazon, USA}
%  \institution{Amazon}
}

\author{Paloma de Juan}
\affiliation{
  \country{Yahoo Research, USA}
%  \institution{Yahoo Research}
}

\begin{comment}
\author{Joao Soares}
\affiliation{
\country{Yahoo Research, USA}
%  \institution{Yahoo Research}
}

\author{Erfan Eshratifar}
\affiliation{
\country{Yahoo Research, USA}
%  \institution{Yahoo Research}
}

\author{Kapil Thadani}
\affiliation{
\country{Yahoo Research, USA}
%  \institution{Yahoo Research}
}
\end{comment}

\begin{abstract}
Online ads showing e-commerce products typically rely on the product images in a catalog sent to the advertising platform by an e-commerce platform.
In the broader ads industry such ads are called dynamic product ads (DPA). It is common for DPA catalogs to be in the scale of millions (corresponding to the scale of products which can be bought from the e-commerce platform). However, not all product images in the catalog may be appealing when directly re-purposed as an ad image, and this may lead to lower click-through rates (CTRs). In particular, products just placed against a solid background may not be as enticing and realistic as a product staged in a natural environment. To address such shortcomings of DPA images at scale, we propose a generative adversarial network (GAN) based approach to generate staged backgrounds for un-staged product images. Generating the entire staged background is a challenging task susceptible to hallucinations. To get around this, we introduce a simpler approach called copy-paste staging using retrieval assisted GANs. In copy paste staging, we first retrieve (from the catalog) staged products similar to the un-staged input product, and then copy-paste the background of the retrieved product in the input image. A GAN based in-painting model is used to fill the holes left after this copy-paste operation. We show the efficacy of our copy-paste staging method via offline metrics, and human evaluation. In addition, we show how our staging approach can enable animations of moving products leading to a video ad from a product image.
\end{abstract}

\maketitle

\section{Introduction} \label{sec:introduction}
\begin{figure}[]%!htb
\centering
  \includegraphics[width=0.8 \columnwidth]{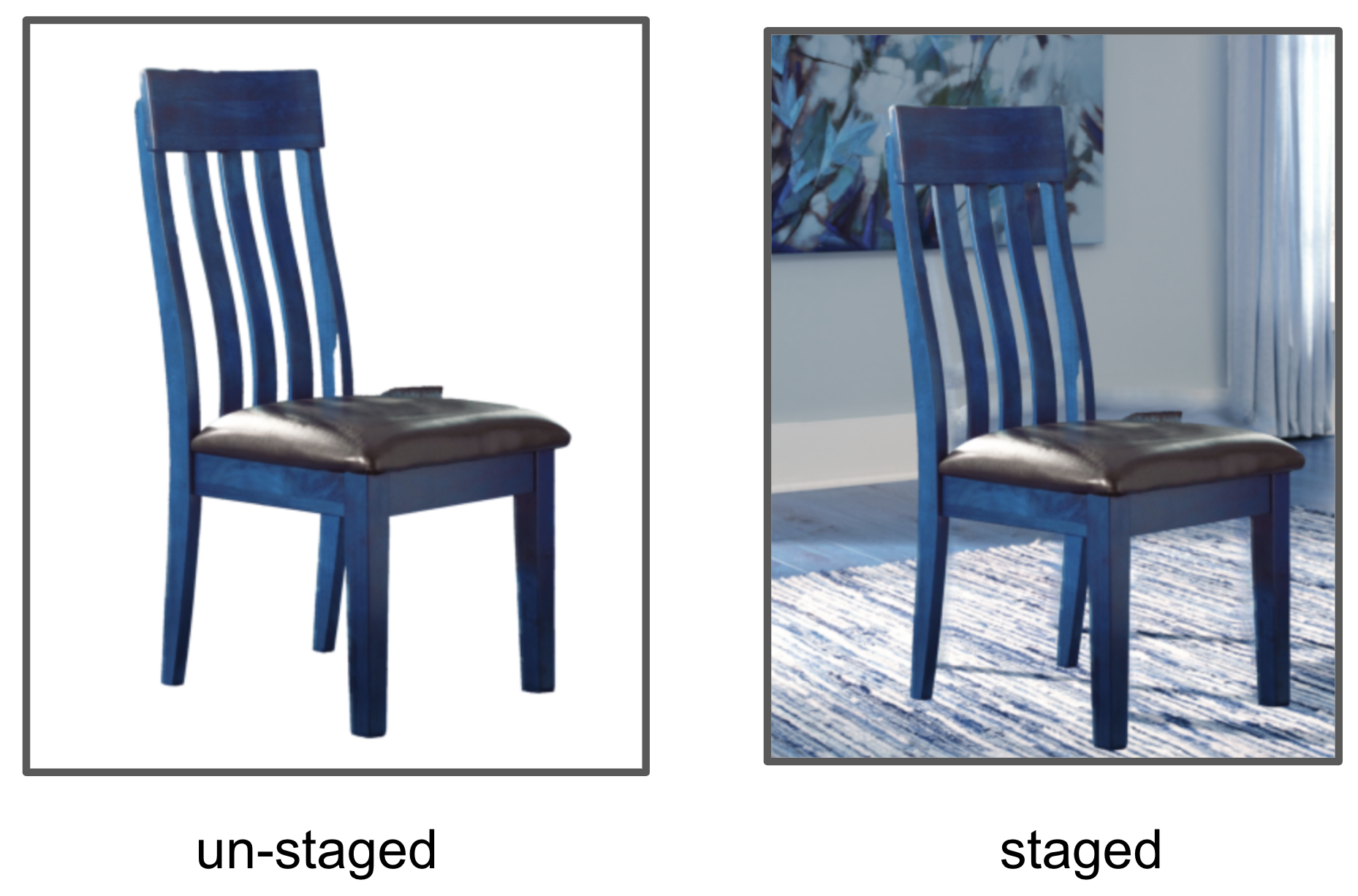}
  \caption{Un-staged and staged version of a chair sold by an e-commerce vendor (staged version is likely to have better CTR).}
  \label{fig:pull_figure}
\end{figure}

The choice of image for an online ad can have a significant impact on the online user exposed to the ad. If the ad image is enticing enough, it can not only create brand awareness among online users but also drive them to click the ad and make subsequent purchases (conversions) \cite{mappi_CIKM,gemx_kdd}. However, if the ad image is not properly designed to capture the user's attention, it would lead to poor user interactions and adversely affect the advertising platform (by lowering revenue) and the advertiser (by lowering conversion rate). In this context, a common observation \cite{cikm2020_createbetterads} is that ad images with products in a natural or real world setting (lifestyle images) tend to have better online performance. For example, an ad selling a chair is expected to perform better if the image shows a chair in a living room versus a chair against a solid (synthetic) background (as shown in Figure~\ref{fig:pull_figure}).
However, such staging of products may be expensive and time consuming, specially when a vendor is selling multiple products at the same time.
In DPA offerings from ad platforms (\emph{e.g.}, Yahoo), the catalog images from an e-commerce vendor (\emph{e.g.}, Walmart, Amazon) are typically used directly as ad images. As described later in our data analysis (based on data from an ad platform), a major fraction of such images are not staged, and hence there is a scope to enhance such images (\emph{e.g.}, by generating a suitable background for the product).

Image generation has been an actively studied topic for the past few years. Diffusion models \cite{dalle} and GANs \cite{goodfellow2014generative,isola2017image} have been powering the state-of-the-art results in this area. Image in-painting \cite{nazeri2019edgeconnect} is a slightly easier version of the problem where only parts of the image need to be generated as opposed to the whole image. To the best of our knowledge, we are the first to study GAN based image generation approaches for enhancing product images to serve as ad images.
Our main contributions can be summarized as follows.
\begin{enumerate}
\item We study three tasks (as outlined below): (i) vanilla staging, (ii) copy-paste staging, and (iii) image-to-parallax animation.
\item In task 1, we aim to generate the entire background for a product. We use pix2pix \cite{isola2017image} to train a GAN model with pairs of images (input: segmented out product image, output: staged product image with ground truth background).
\item In task 2, our goal is to retrieve a similar product image (with staging) and copy-paste the background while filling in gaps (holes) created in the process of swapping products. We leverage GAN-based in-painting to fill in the gaps mentioned above. We also introduce a weighted boundary loss for in-painting to focus on the image generation quality at product boundaries. Through Frechet inception distance (FID) score, and human evaluations we show that copy-paste staging is significantly better than the vanilla staging baseline.

\item In task 3, we use GAN-based in-painting to create a sequence of images simulating the main product's movement against the staged background as in a parallax animation. The foreground and background both move, but at different speeds, creating the illusion of depth. This is to show how our approach can lead to video ads from product images.
\end{enumerate}
Our retrieval based approach (second task above) shares the intuition common in text generation: retrieval augmented generation (RAG) has better context understanding and generation quality. The remainder of this paper is organized as follows: related work in Section~\ref{sec:related}, problem formulation in Section~\ref{sec:formulation}, relevant data in Section~\ref{sec:data}, and our proposed approaches in Section~\ref{sec:method}. We go over our experimental results in Section~\ref{sec:results}, and end with a discussion in Section~\ref{sec:discussion}.
\section{Related work} \label{sec:related}
\paragraph*{Online advertising:} In online advertising, the ad creative (text and image) plays an important role in influencing online users towards brand awareness, clicks and purchases \cite{gemx_kdd,visualtextrank,recsys_2023_shaunak}. Studying ad images and text using state-of-the-art deep learning models in computer vision and natural language processing (NLP) is an emerging area of research. In \cite{cvpr_kovashka}, ad image content was studied using computer vision models, and their dataset had manual annotations for: ad category, reasons to buy products advertised in the ad, and expected user response given the ad.
%However, \emph{understanding ad creatives from a brand's perspective} was missing in both \cite{kovashka_eccv2018, cvpr_kovashka}, and 
Using this dataset, \cite{self_recsys2019, www20_joey} used ranking models to recommend themes for ad creative design using a brand's Wikipedia page. In \cite{cikm2020_createbetterads}, object tag recommendations for improving an ad image was studied using data from A/B tests. Although related to ads, the above methods are not applicable in our setup since none of them are image generation methods.

\paragraph*{GANs for image generation}
Generative Adversarial Networks (GANs) are a popular approach for image generation. While vanilla GANs~\cite{goodfellow2014generative} generate images from random noise, Conditional GANs~\cite{mirza2014conditional} also allow to use extra information in a generative process. Recently developed \verb|pix2pix| method~\cite{isola2017image} and its successors~\cite{wang2018high} use Conditional GANs for image-to-image generation optimizing GAN objective together with distance to a target image. While we focus on GANs in this paper, our copy-paste staging approach can be generalized to more recent diffusion models \cite{dalle} (discussed in Section~\ref{sec:discussion}).

\paragraph*{Saliency detection}
Saliency detection in product images is needed to understand which parts of an image correspond to the main product being advertised (as opposed to the background). Salient object detection (SOD) aims to detect the most visually attractive objects with precise boundaries in images (\emph{i.e.}, it returns a boundary map which can be used to segment out objects from the image). With the introduction of convolutional neural networks (CNNs) in computer vision, SOD accuracy has witnessed remarkable improvements. Recently, U\textsuperscript{2}-Net \cite{qin2020u2} achieved state-of-the-art results for saliency detection by using a nested version of U-Net \cite{ronneberger2015u}-like architecture to capture richer local and global information. In our proposed approaches, we use saliency detection via U\textsuperscript{2}-Net as a building block.

\section{Problem formulation} \label{sec:formulation}
\begin{figure}[]%!htb
\centering
  \includegraphics[width=1 \columnwidth]{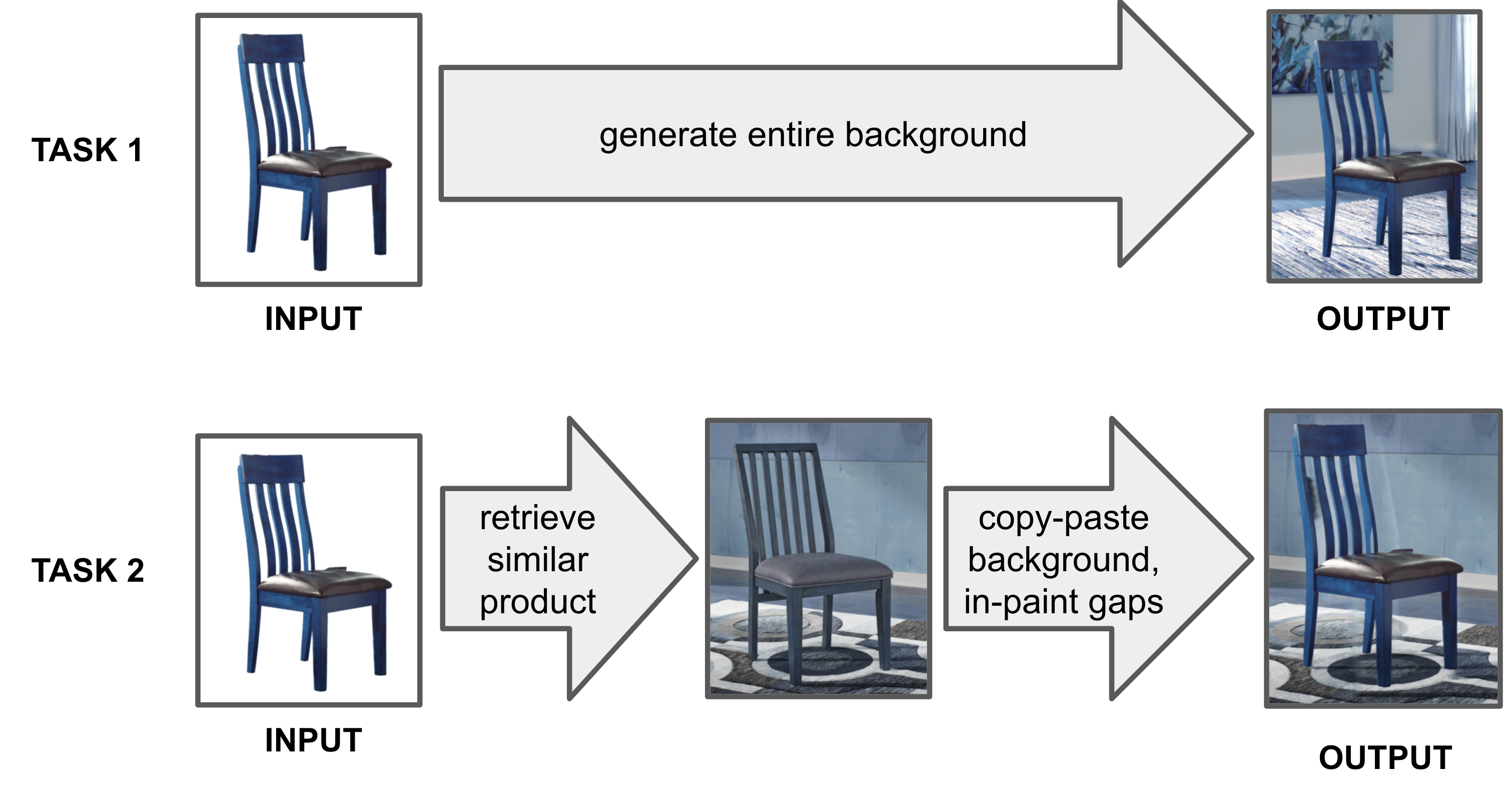}
  \caption{Tasks 1 and 2 explained with examples. In task 2, we need to retrieve a related product image with staging and then copy paste its background onto the input image, whereas in task 1 the goal is to generate the entire background.}
  \label{fig:tasks1_2}
\end{figure}

We study three tasks in this paper as outlined below.
\paragraph*{Task 1: vanilla staging.}
In task 1, the input is a product image (without staging), and the desired output is a product image with a model generated \textit{relevant} background (stage). The model should generate the entire background as shown in Figure~\ref{fig:tasks1_2}.
\paragraph*{Task 2: retrieval assisted copy-paste staging.}
Task 2 is a simpler version of task 1. Here, we are given a pool of existing product images $\mathcal{P}$, and we need to retrieve a similar product image with staging such that we can copy-paste the staged background from the retrieved image onto the input image as shown in Figure~\ref{fig:tasks1_2}. Image generation (in-painting) is needed to fill in the gaps after copy-pasting (since the input product and the product in the retrieved image are not identical, gaps will be created when we swap products).

\paragraph*{Task 3: image-to-parallax animation.}
In this task, the goal is to take an input image (as shown in Figure~\ref{fig:tasks1_2} for task 2), and create an animation (\emph{i.e.}, sequence of images), where the object in the input image (as in Figure \ref{fig:tasks1_2}) appears to be moving against a stationary but staged background. Such animations are expected to lead to higher user engagement \cite{verizon_media_interactive_ads_report}.

\section{Data} \label{sec:data}
\begin{figure}[]%!htb
\centering
  \includegraphics[width=0.7 \columnwidth]{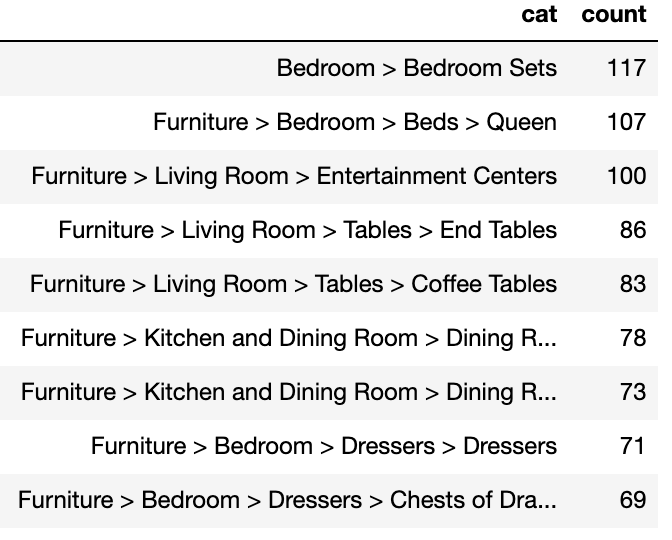}
  \caption{Subcategories for the ``Furniture" category and counts.}
  \label{fig:furn_subcat}
\end{figure}

For our experiments, we sampled data from Yahoo Gemini DPA (spanning November-December 2020). With an impression threshold of $10,000$, we collected a sample of $\sim 18,899$ ads (each corresponding to a product), from which $11,927$ had solid backgrounds, and the rest $6,972$ had staged backgrounds. Each product in the sample belongs to a certain hierarchical category, e.g. "Kids $>$ Kids Nightstands", "Furniture $>$ Bedroom $>$ Headboards $>$ Queen", "Shoe $>$ Heel $>$ Oxford Heel". For our experiments, we considered only furniture images with staged backgrounds, which left us with $2071$ images. Top-frequent subcategories with instance counts for the ``Furniture" category (as in our sample) are shown in Figure~\ref{fig:furn_subcat}. 

\begin{comment}
\subsection{Public dataset: DUTS}
%\subsubsection{TBD. Amazon, and MIT?}
We also use the DUTS image dataset \cite{wang2017learning} in our experiments. DUTS provides 10,553 training images and 5,019 testing images, containing challenging saliency detection scenarios. Specifically, we look up the synset ID of WordNet and map the DUTS images to different categories, and mainly focus on images in furniture category.This dataset is used for training the GAN models used in our proposed approaches as described in Section \ref{sec:method}.
\end{comment}

\section{Product staging using GANs} \label{sec:method}
In this section, we first explain the salient object detection model which is common to our proposed approaches for all the tasks outlined in Section~\ref{sec:formulation}.
Next, we cover our proposed approaches for tasks $1$, $2$, and $3$ (tasks explained in Section~\ref{sec:formulation}) in Sections~\ref{sec:method_task_1}, \ref{sec:method_task_2}, and \ref{sec:method_task_3} respectively. Our major modeling contributions are in the approaches for tasks 2 and 3; for task 1, although we define the task, we use existing approaches (pix2pix \cite{isola2017image}) to solve it. Task 2 is an easier version of task 1, and our proposed approach leads to more realistic staged product images compared to pix2pix for task 1.

\subsection{Product segmentation via saliency} \label{sec:method_saliency}
For tasks $1$, $2$ and $3$, we use U\textsuperscript{2}-Net \cite{qin2020u2} as our saliency object detector. The saliency object detector plays a crucial role in the first step of our approaches since it separates the main product(s) that will be replaced or copy-pasted versus the background in a product image. Once saliency probability maps are obtained from U\textsuperscript{2}-Net, we set the threshold at $0.5$ to generate binary masks and separate foreground pixels from the background pixels. For each task, we use product segmentation in a different manner as explained below.

\subsection{Vanilla staging} \label{sec:method_task_1}
\begin{figure}[]%!htb
\centering
  \includegraphics[width=0.8 \columnwidth]{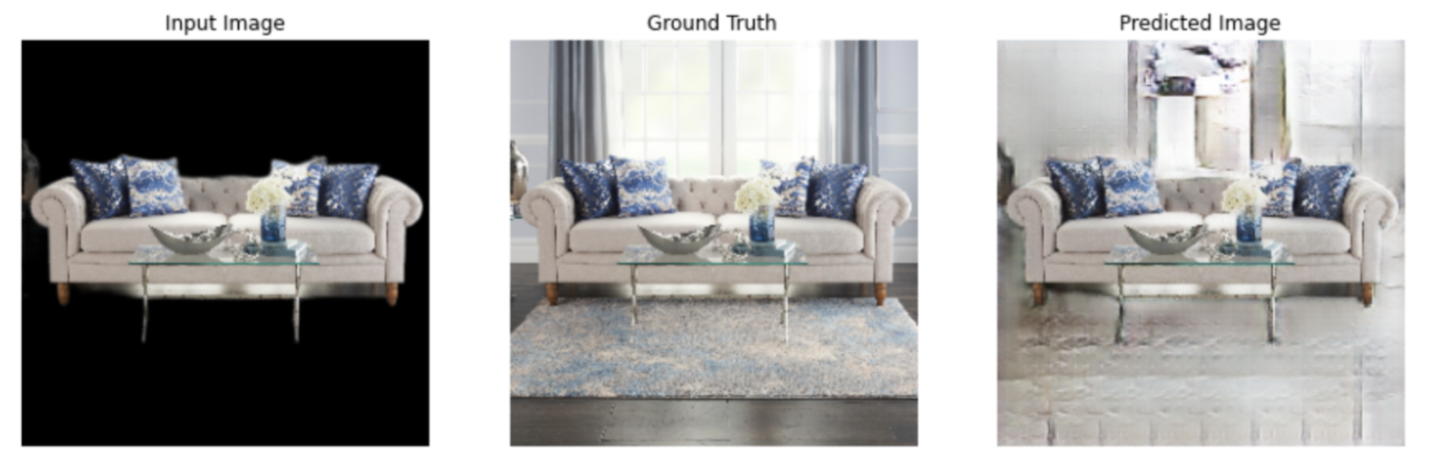}
  \caption{Example of vanilla staging using pix2pix.}
  \label{fig:pix2pix_ex}
\end{figure}

For vanilla staging, we use the \verb|pix2pix| method to generate an image background for a product which is segmented by a saliency mask. The algorithm is a conditional GAN optimizing the loss combining 1) regular GAN objective, and 2) $\ell_1$ distance between original and restored images.  We use product segmentation to prepare pairs of images to train pix2pix for stage (background generation). In particular, given an image with a staged product, we remove the background via product segmentation and use this as the input image to pix2pix. Figure~\ref{fig:pix2pix_ex} shows an example for this approach (original image in the middle, segmented product on the left, and the version with generated staging on the right). 

\subsection{Retrieval assisted copy-paste staging}  \label{sec:method_task_2}
\begin{figure}[]%!htb
\centering
  \includegraphics[width= 1 \columnwidth]{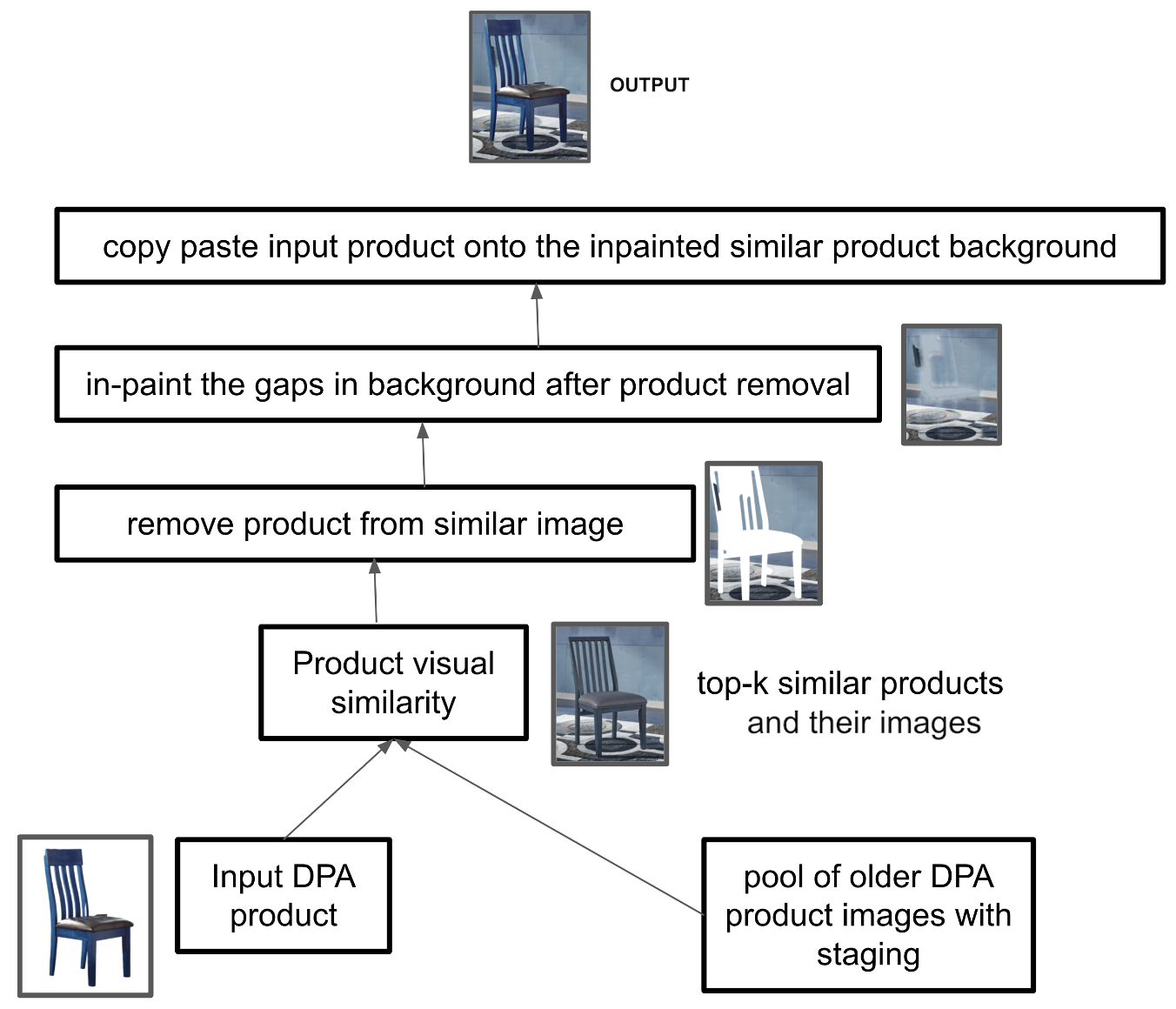}
  \caption{Proposed copy-paste staging algorithm.}
  \label{fig:cp_flow}
\end{figure}

\begin{figure}
\begin{subfigure}{\columnwidth}
  \centering
  \includegraphics[width=.6\linewidth]{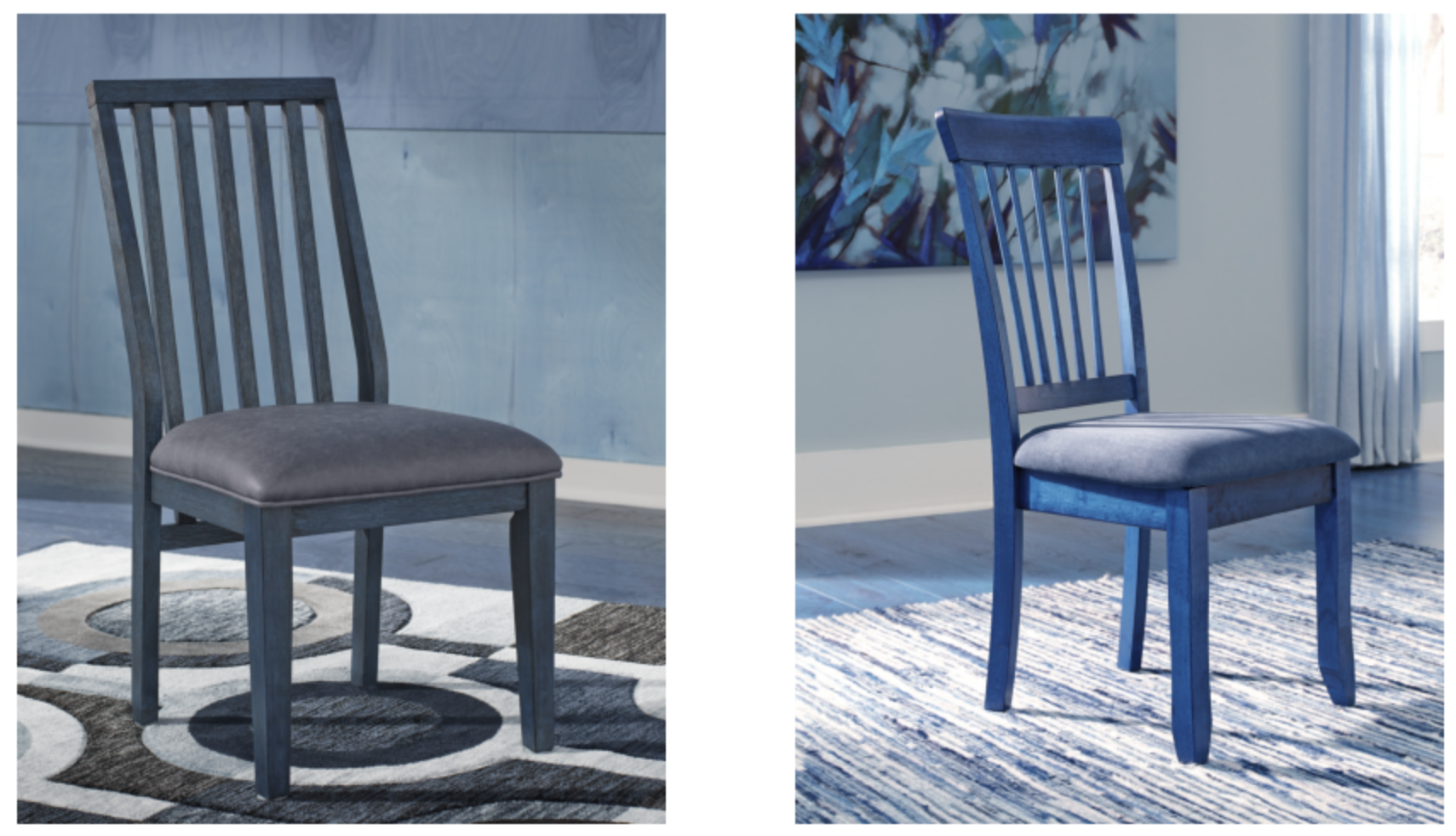}
  \caption{Top-2 similar images}
  \label{fig:top_2_sim}
\end{subfigure} \\
\begin{subfigure}{\columnwidth}
  \centering
  \includegraphics[width=.6\linewidth]{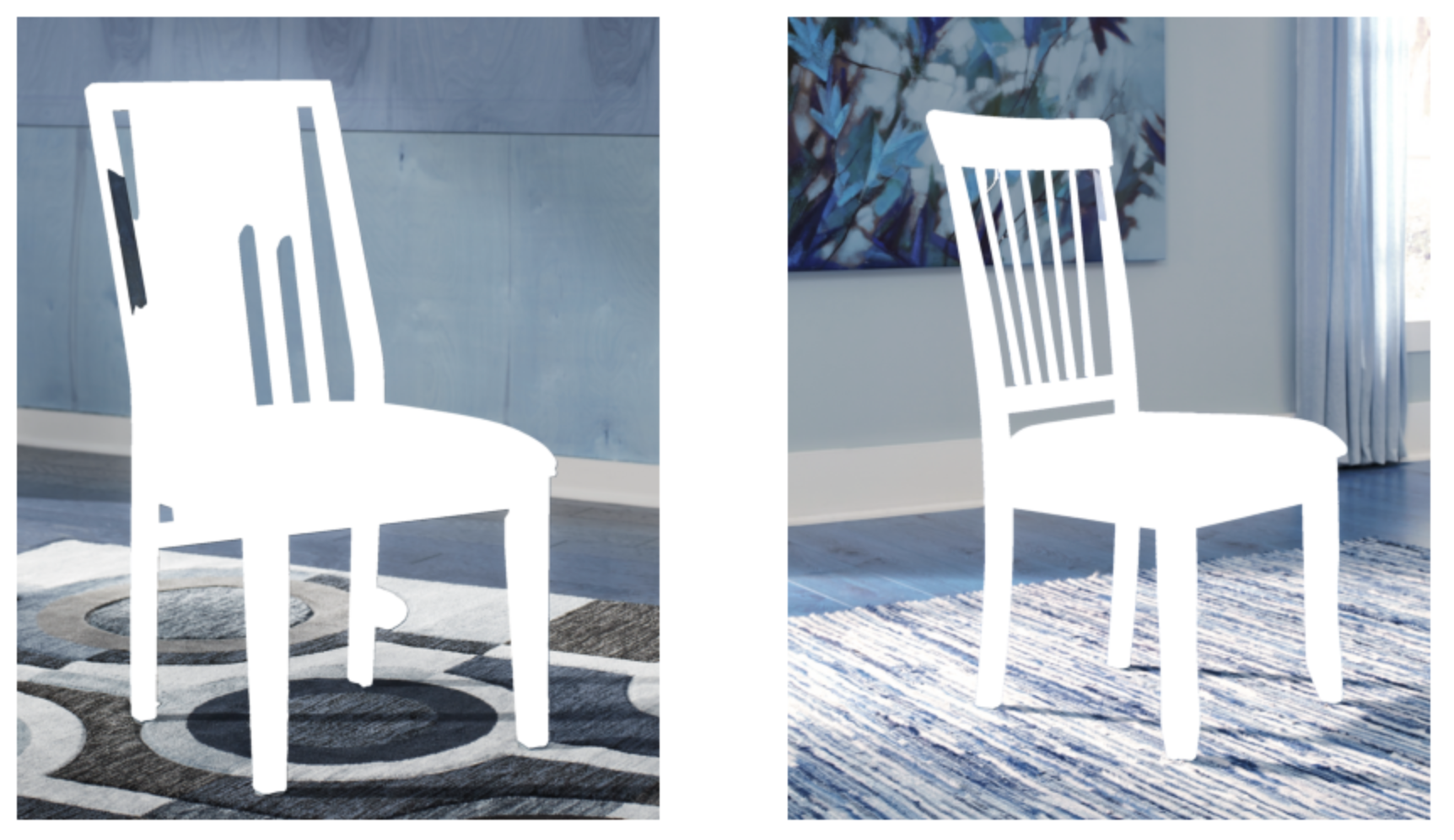}
  \caption{Products masked out}
  \label{fig:top_2_masked_out}
\end{subfigure} \\
\begin{subfigure}{\columnwidth}
  \centering
  \includegraphics[width=.6\linewidth]{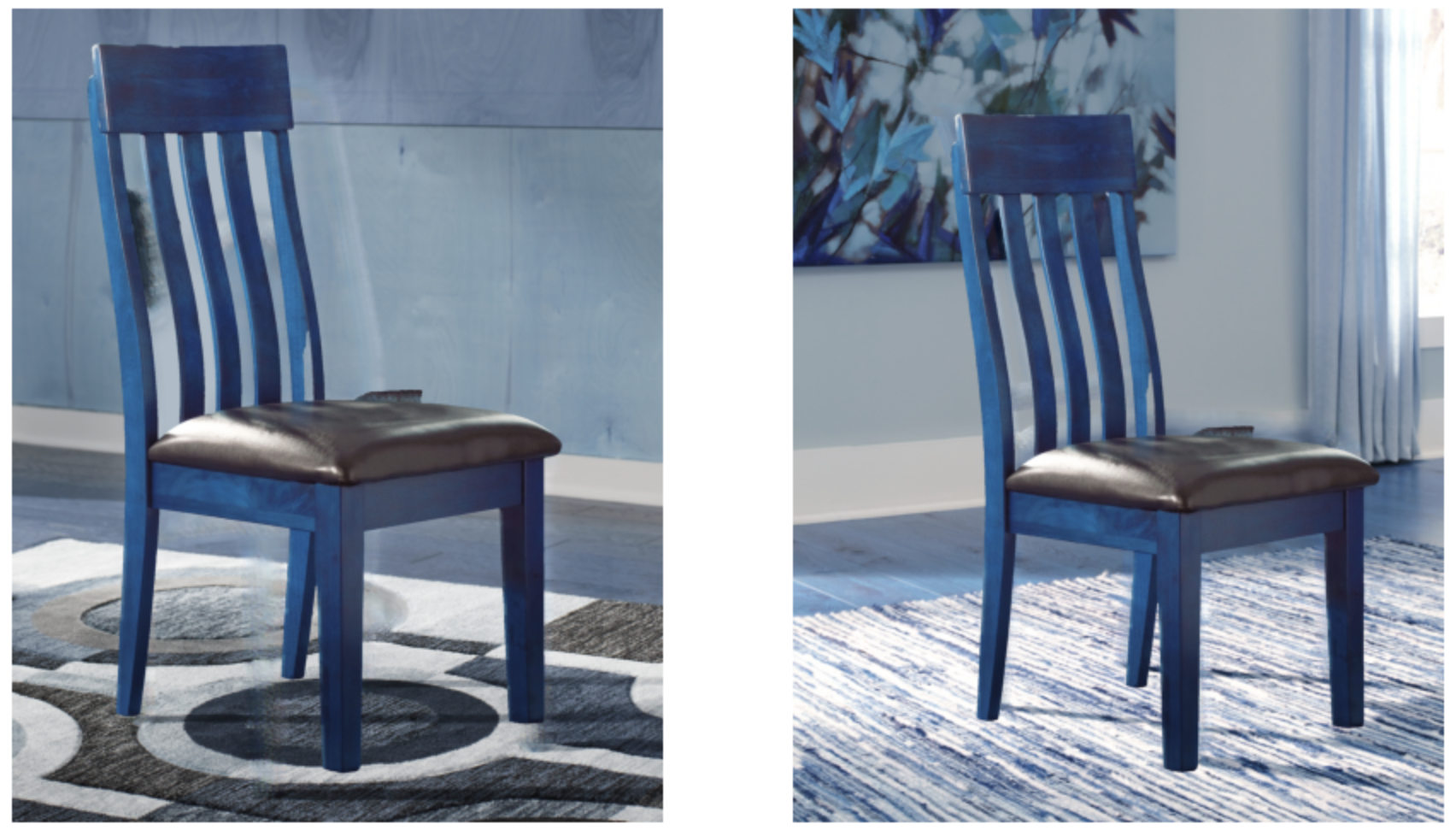}
  \caption{Copy-paste result after inpainting}
  \label{fig:top_2_res}
\end{subfigure}
\caption{End-to-end result of copy-paste staging for top-2 similar images} \label{fig:cp_example_0}
\end{figure}

For copy paste staging, we bypass the problem of generating the entire image background by using backgrounds from other relevant images. Our method consists of the following steps:
\begin{enumerate}
    \item For a given segmented product (Figure~\ref{fig:pull_figure}, left), retrieve top-$k$ similar products from a training collection (Fig.~\ref{fig:top_2_sim}). Similarity measure is a cosine distance between embeddings of corresponding product images provided by \verb|Inception-V3|~\cite{szegedy2016rethinking}.
    \item For the top-$k$ similar images, segment out the original products (Fig.~\ref{fig:top_2_masked_out}) and fill in the holes by inpainting using EdgeConnect \cite{nazeri2019edgeconnect} (GAN based model) and a new loss function that we introduce in Section~\ref{sec:wbl}.
    \item Copy-paste the original product image to the inpainted top-$k$ similar images, aligning shape and center mass for the corresponding product masks (Fig.~\ref{fig:top_2_res}). 
\end{enumerate}
The above algorithm is illustrated with examples in Figure \ref{fig:cp_flow}. We provide additional examples in Figure~\ref{fig:cp_example_0}.
After completing steps 1-3, we generate $k$ product images with various backgrounds, only small parts of which (holes around the product before/after) are generated by GANs, which makes the images look more real if comparing with vanilla staging. For better background generation we introduce a new loss function as described below.

\subsubsection{Weighted Boundary Loss} \label{sec:wbl}
\begin{figure}
     \centering
     \begin{subfigure}[h]{0.32\linewidth}
         \centering
         \includegraphics[width=1\linewidth]{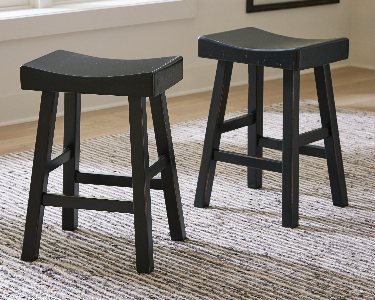}
         \caption{}
         \label{fig:loss_input}
     \end{subfigure}
     \hfill
     \begin{subfigure}[h]{0.32\linewidth}
         \centering
         \includegraphics[width=1\linewidth]{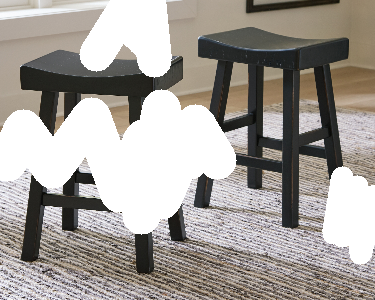}
         \caption{}
         \label{fig:loss_masked}
     \end{subfigure}
     \hfill
     \begin{subfigure}[h]{0.32\linewidth}
         \centering
         \includegraphics[width=1\linewidth]{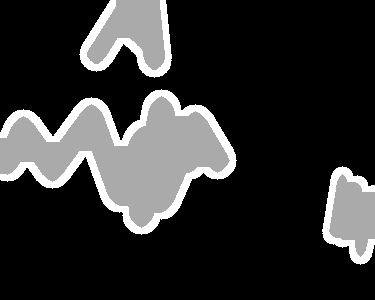}
         \caption{}
         \label{fig:loss_weighted_mask}
     \end{subfigure}
        \caption{Illustration of weighted boundary loss. \ref{fig:loss_input}: Ground-truth training data. \ref{fig:loss_masked}: Free-form masked training data. \ref{fig:loss_weighted_mask}: Weighted mask.}
        \label{fig:loss_mask_example}
\end{figure}

Recent works \cite{nazeri2019edgeconnect} and \cite{wang2021image} explore coarse-to-fine inpainting approaches, since the structures of objects are complex and diverse, adding an intermediate step, like edge maps or monochromic images, can help models to learn progressively and eventually generate better final inpainted outputs. We propose a weighted boundary loss (WBL) to not only simplify the learning process (since the model needs to focus on lesser area), but also mimic the end application use case.  Following prior work \cite{nazeri2019edgeconnect}, our total generator loss consists of a conventional adversarial loss $L_{ADV}$ and a feature-matching loss $L_{FM}$. In addition to these two losses, since our goal is to make the model learn better at the boundary of the masked area, we add weighted boundary loss $L_{WBL}$ to amplify the loss penalty at the boundary area pixels. WBL is: \begin{equation}
L_{WBL}=W_{map}*L_{\ell_1-norm}(E_{GT}, E_{pred}),
\end{equation}
where $E_{GT}$ is ground truth edge map of input images, $E_{pred}$ is predicted edge map generated by the generator. The $W_{map}$ is a pixel-wise weighted map and has the same size as input masked images and ground truth. To be more specific, the $W_{map}$ has $\lambda_{boundary}$ for pixels around the boundary between masked area and unmasked area, and $\lambda_{non-boundary}$ for pixels away from the boundary, the pixel-wise $L_{l1-norm}$ will multiply the corresponding $\lambda$ as we calculate $L_{WBL}$. As Figure~\ref{fig:loss_mask_example} illustrates, for each training sample, we create free-form dense masks by the method proposed by \cite{yu2019free}. Then, we find the boundary area of the free-form mask and assign $\lambda_{boundary}$ (white area in Figure~\ref{fig:loss_weighted_mask}) and $\lambda_{non-boundary}$ (gray area in Figure~\ref{fig:loss_weighted_mask}). For experiments, we fixed $\lambda_{boundary}=0.9$ and $\lambda_{non-boundary}=0.1$.

\subsection{Image-to-animation}  \label{sec:method_task_3}
\begin{figure}
     \centering
     \begin{subfigure}[h]{0.3\linewidth}
         \centering
         \includegraphics[width=0.9\linewidth]{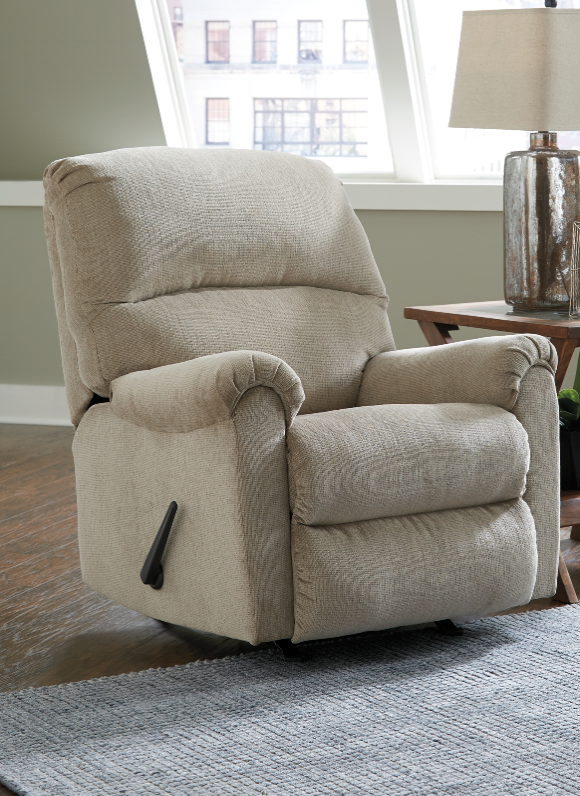}
         \caption{}
         \label{fig:parallax_a}
     \end{subfigure}
     \hfill
     \begin{subfigure}[h]{0.3\linewidth}
         \centering
         \includegraphics[width=0.9\linewidth]{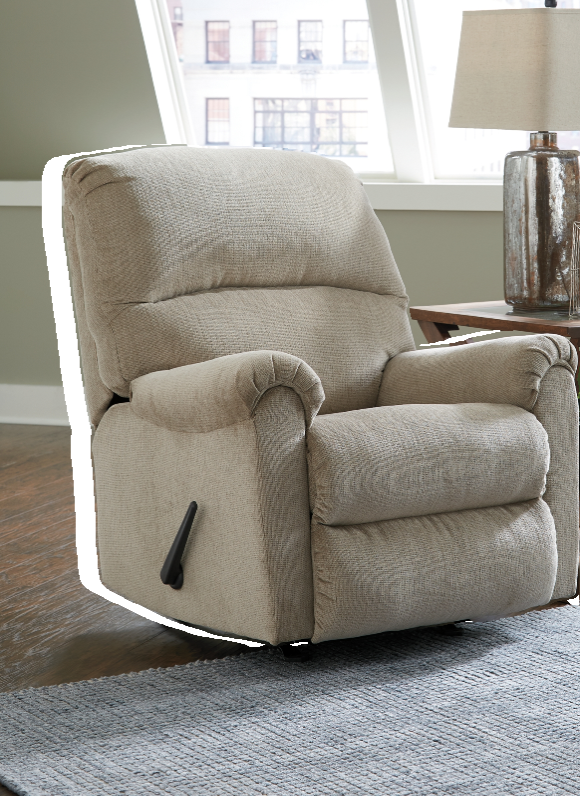}
         \caption{}
         \label{fig:parallax_b}
     \end{subfigure}
     \hfill
     \begin{subfigure}[h]{0.3\linewidth}
         \centering
         \includegraphics[width=0.9\linewidth]{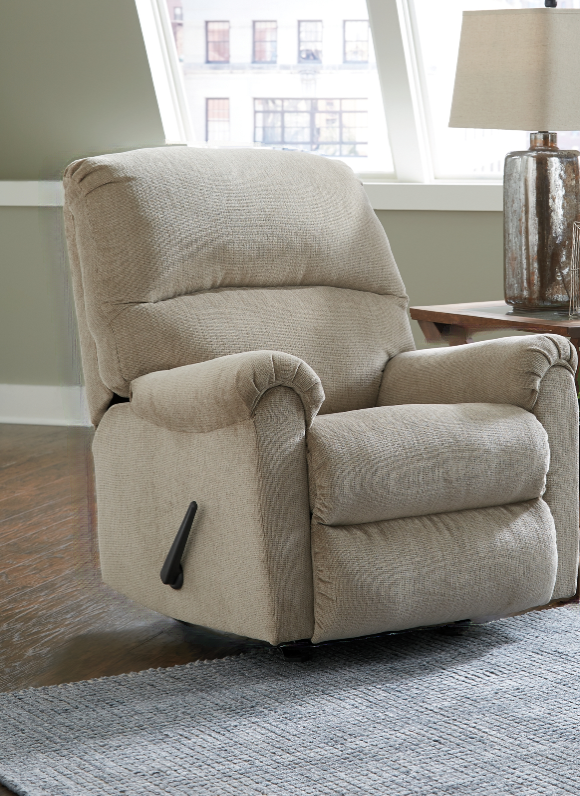}
         \caption{}
         \label{fig:parallax_c}
     \end{subfigure}
        \caption{Illustration of image parallax effect. \ref{fig:parallax_a}: original image \ref{fig:parallax_b}: image after shifting all the salient pixels. \ref{fig:parallax_c}: gap in-painted.}
        \label{fig:parallax_example}
\end{figure}

Parallax effect happens when the background pixels move slower than foreground objects in an animation, thereby creating an illusion of depth in a two-dimensional image. Generally, parallax effect requires independent foreground images and background images, and proper technique to make transparent backgrounds. In our proposed approach, by leveraging the power of salient object detection and in-painting, a parallax effect animation can be generated from a 2D image. Practically, we run salient object detection to define foreground pixels, then gradually move the foreground object around creating empty gaps between the current position of the object and original position. To fill the empty gap, we then use image in-painting model to in-paint those pixels and create serial realistic images. We illustrate the sample results of the above approach in Figure~\ref{fig:parallax_example}.

\section{Results}\label{sec:results}
We first go over some sample results for tasks 1-3 followed by offline metrics (retrieval performance, generation quality) and human perceptual study results.

\subsection{Sample results for tasks 1, 2 and 3}

\begin{figure}
\begin{subfigure}{\columnwidth}
  \centering
  \includegraphics[width=.8\linewidth]{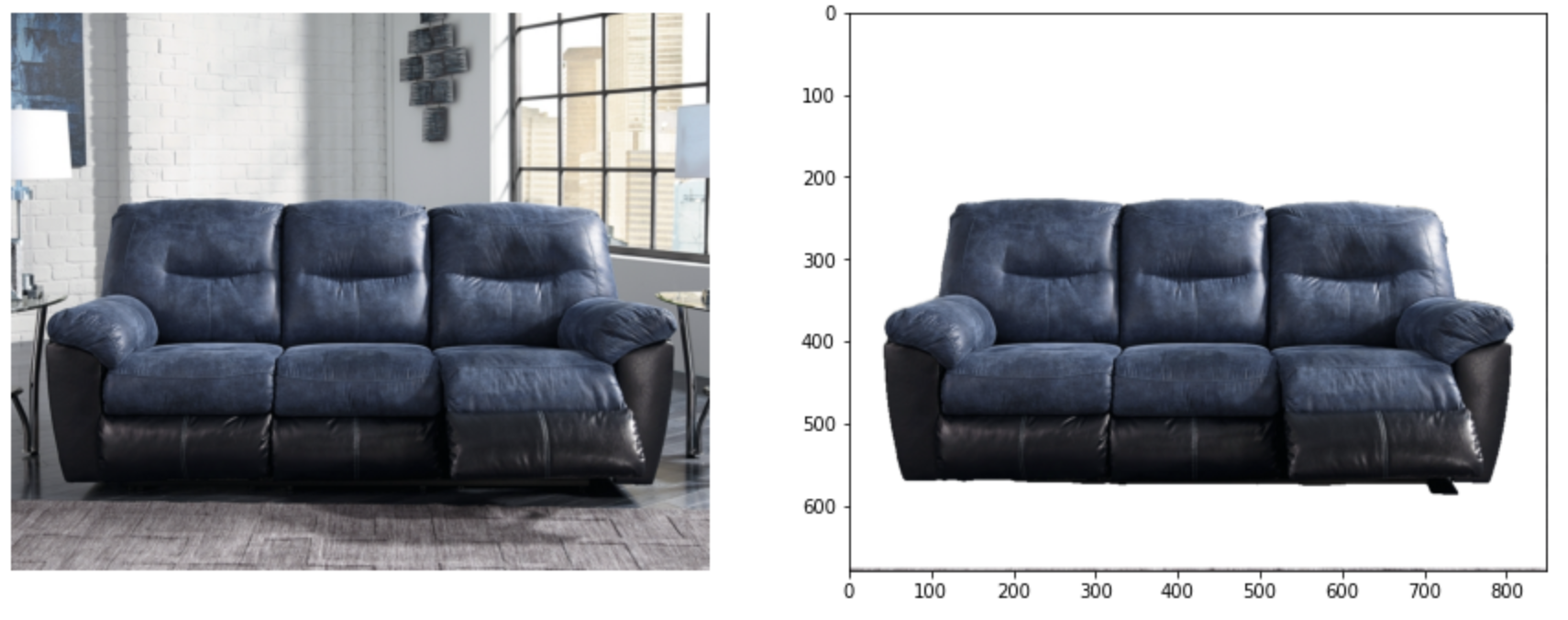}
  \caption{Original image from DPA catalogs (left) and the input product image for staging (right).}
  \label{fig1:orig}
\end{subfigure} \\
\begin{subfigure}{\columnwidth}
  \centering
  \includegraphics[width=.5\linewidth]{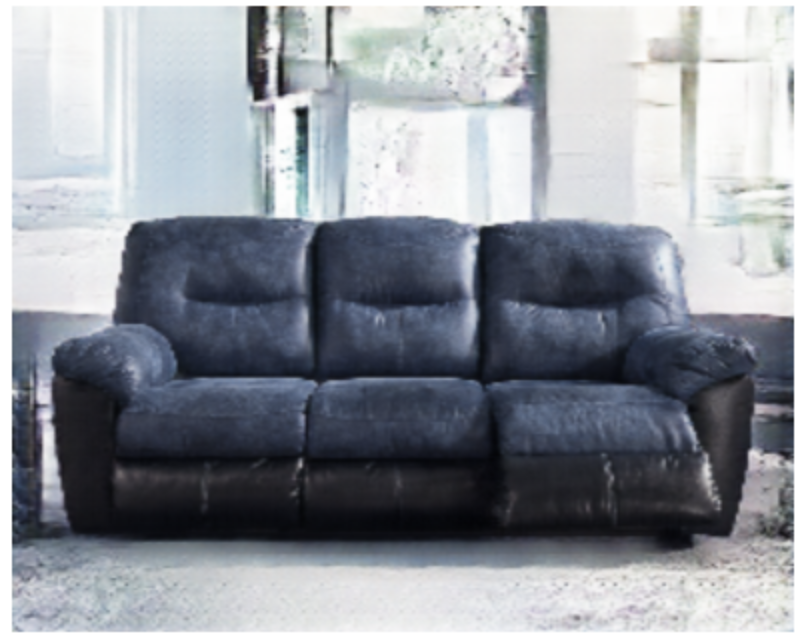}
  \caption{Result of vanilla staging (pix2pix)}
  \label{fig1:vanilla}
\end{subfigure} \\
\begin{subfigure}{\columnwidth}
  \centering
  \includegraphics[width=.8\linewidth]{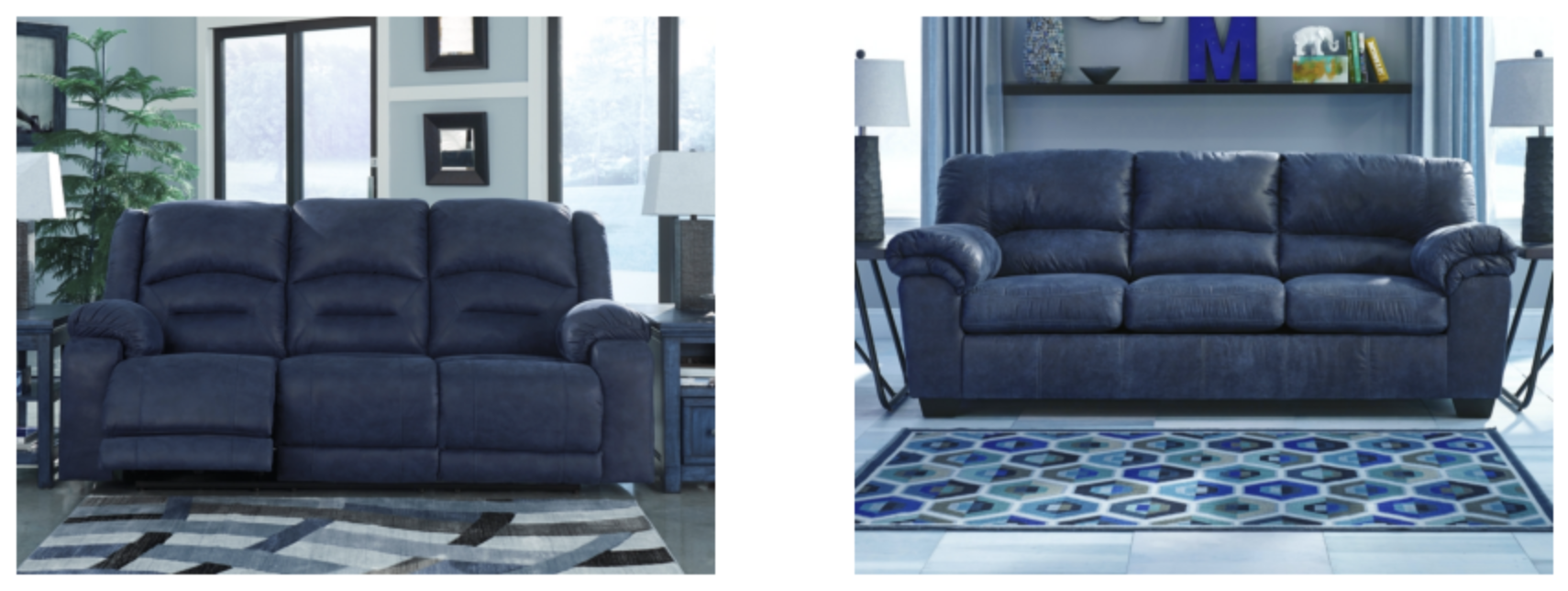}
  \caption{Top-2 similar images}
  \label{fig1:top_2_res}
\end{subfigure} \\
\begin{subfigure}{\columnwidth}
  \centering
  \includegraphics[width=.8\linewidth]{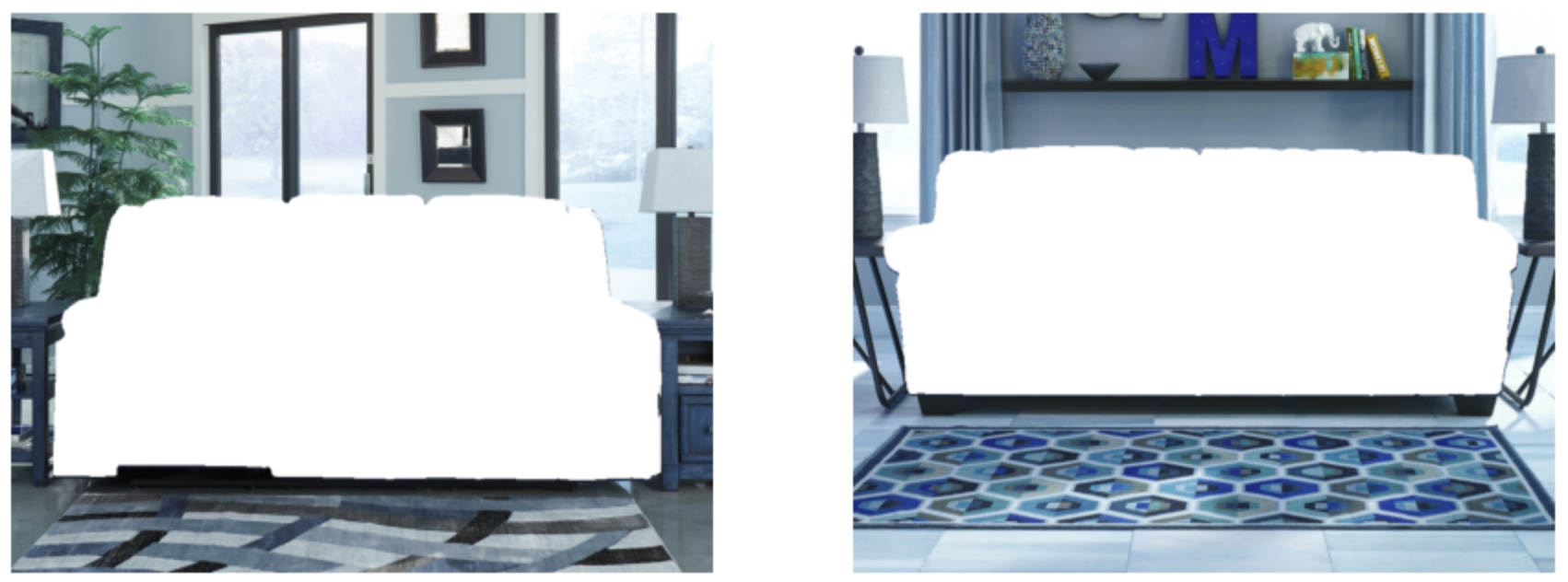}
  \caption{Products masked out}
  \label{fig1:top_2_masked}
\end{subfigure} \\
\begin{subfigure}{\columnwidth}
  \centering
  \includegraphics[width=.8\linewidth]{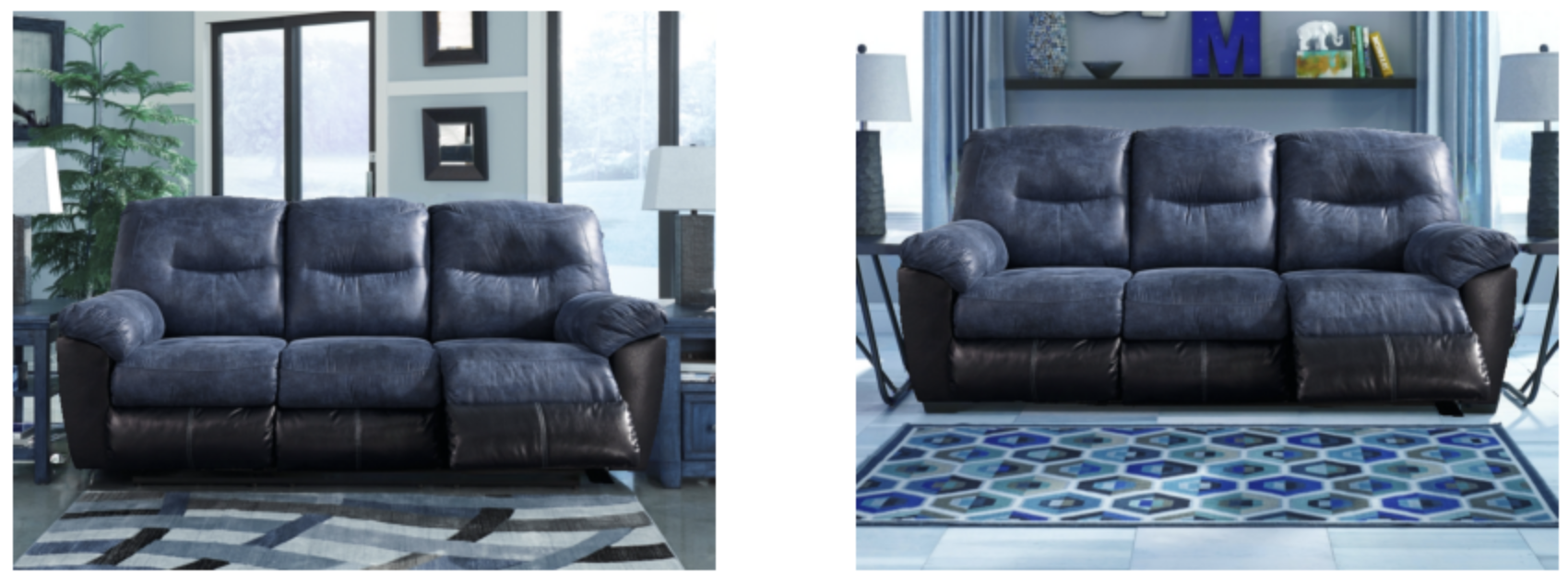}
  \caption{Results of Copy-paste staging}
  \label{fig1:copy_paste_res}
\end{subfigure}
\caption{Example 1 for copy-paste staging.} \label{fig:cp_example_1}
\end{figure}

\begin{figure}
\begin{subfigure}{\columnwidth}
  \centering
  \includegraphics[width=.8\linewidth]{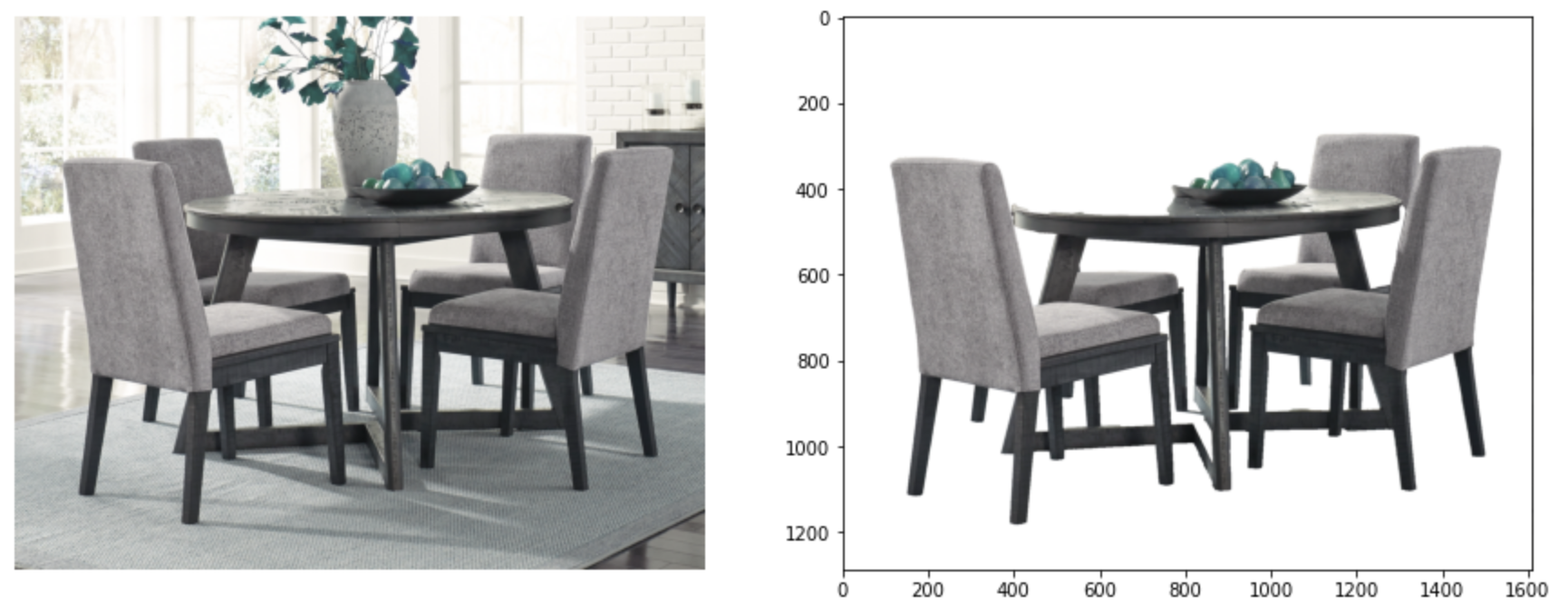}
  \caption{Original image from DPA catalogs (left) and the input product image for staging (right).}
  \label{fig2:orig}
\end{subfigure} \\
\begin{subfigure}{\columnwidth}
  \centering
  \includegraphics[width=.5\linewidth]{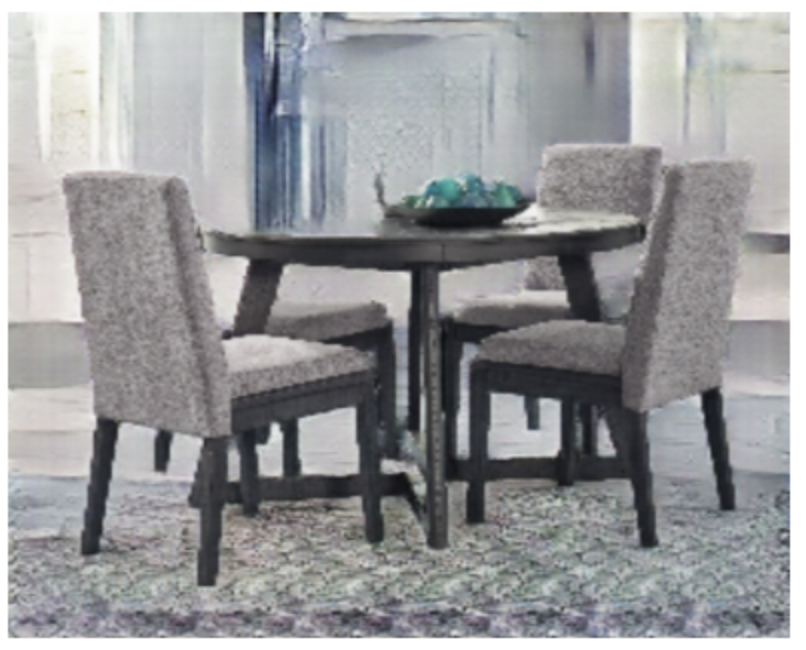}
  \caption{Result of vanilla staging (pix2pix)}
  \label{fig2:vanilla}
\end{subfigure} \\
\begin{subfigure}{\columnwidth}
  \centering
  \includegraphics[width=.8\linewidth]{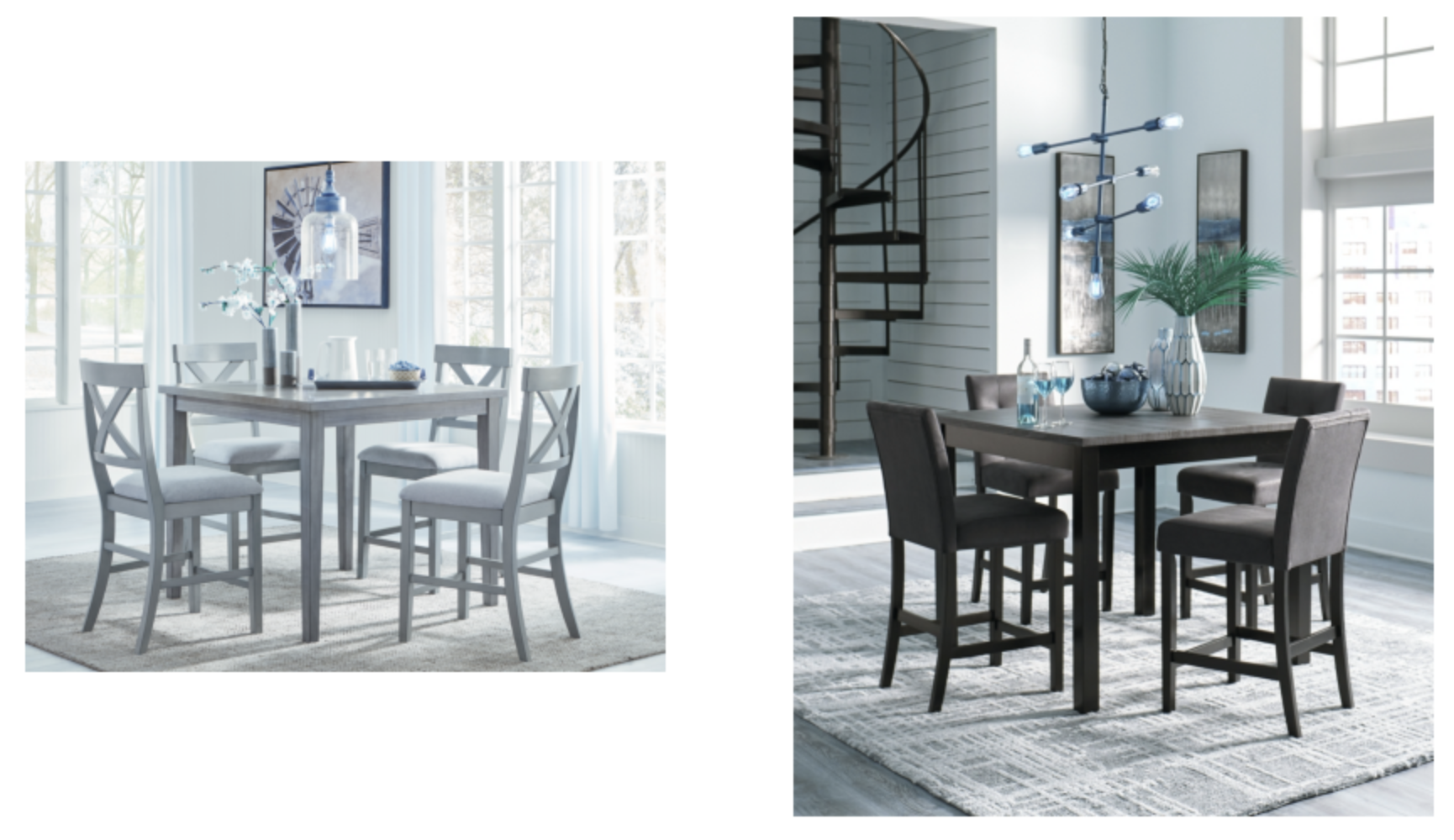}
  \caption{Top-2 similar images}
  \label{fig2:top_2_res}
\end{subfigure} \\
\begin{subfigure}{\columnwidth}
  \centering
  \includegraphics[width=.8\linewidth]{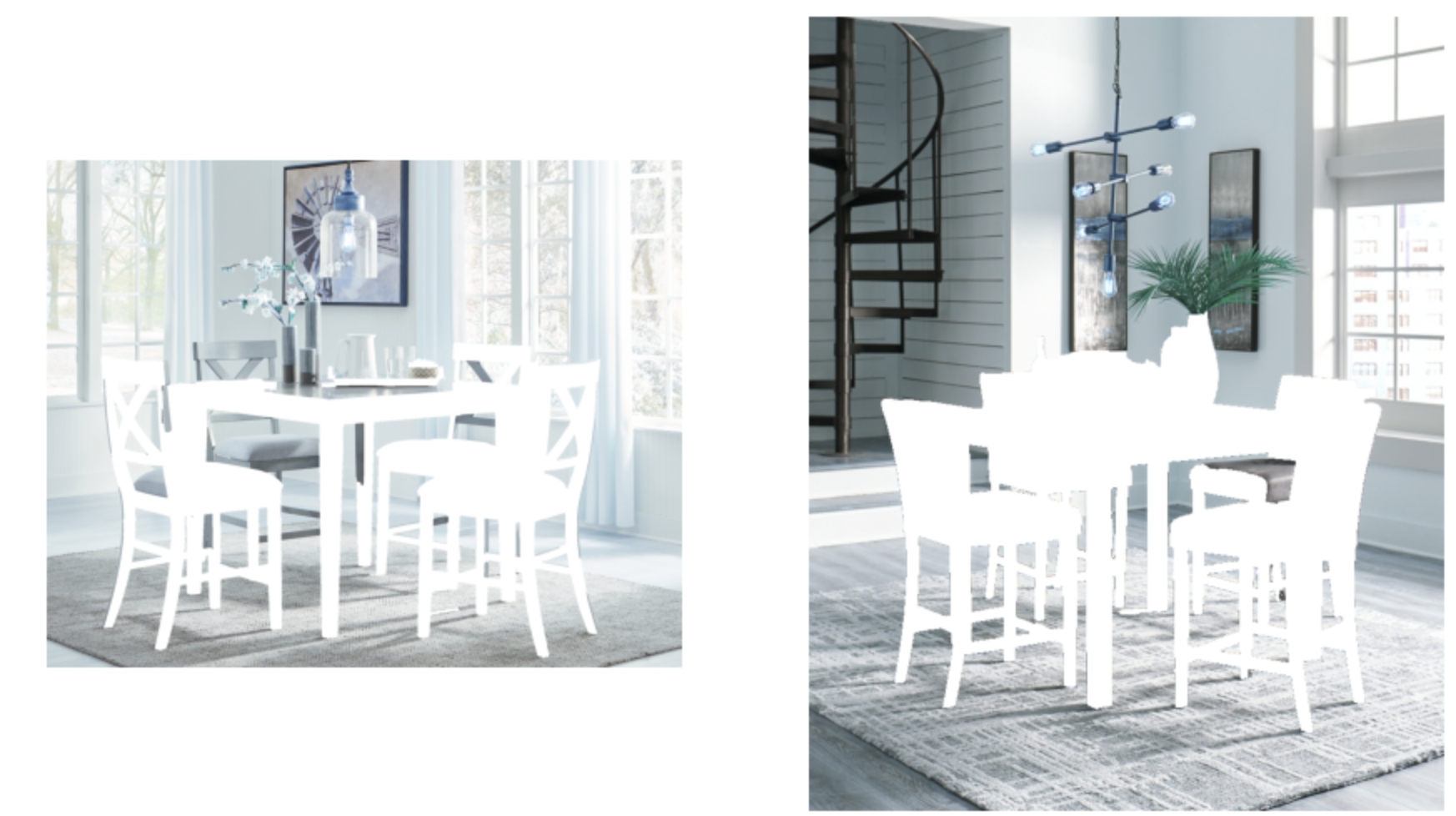}
  \caption{Products masked out}
  \label{fig2:top_2_masked}
\end{subfigure} \\
\begin{subfigure}{\columnwidth}
  \centering
  \includegraphics[width=.8\linewidth]{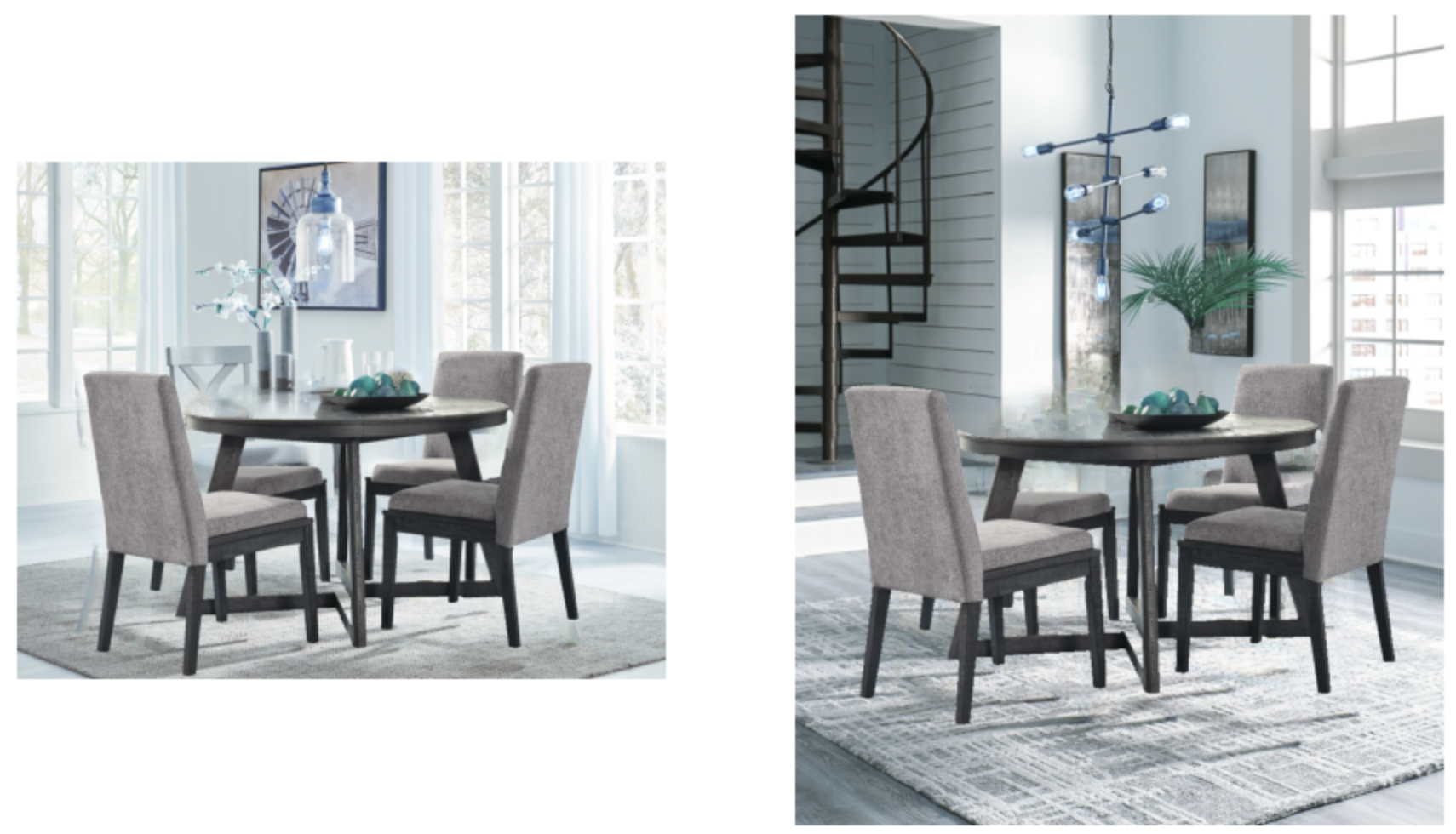}
  \caption{Results of Copy-paste staging}
  \label{fig2:copy_paste_res}
\end{subfigure}
\caption{Example 2 for copy-paste staging.} \label{fig:cp_example_2}
\end{figure}

\paragraph*{Task 1 (vanilla staging):} sample results for task 1 (obtained via pix2pix) are shown in Figures~\ref{fig:cp_example_1} (b) and \ref{fig:cp_example_2} (b). Compared to the original image with the background, the generated image has a lot of artifacts, and does not look so realistic. As we discuss below, the copy-paste staging results look more realistic.

\paragraph*{Task 2 (copy-paste staging):} Figures~\ref{fig:cp_example_1} (e) and \ref{fig:cp_example_2} (e) show sample results for task 2 using the proposed copy-paste staging approach. Overall, the copy-paste staging results look much more realistic compared to pix2pix results.

\paragraph*{Task 3 (parallax animation)} A sample for the generated image-to-parallax animation: \url{https://www.dropbox.com/s/9at5gz24ukhf2gi/product_staging_image_to_parallax_demo.mp4?dl=0}.

\subsection{Offline metrics}
\paragraph*{Similar product image retrieval performance:} a retrieved image is considered similar, if it belongs to the same subcategory as the input product. For example, if the input is a queen bed of subcategory "Furniture $>$ Bedroom $>$ Headboards $>$ Queen", the retrieved product should fall into the same subcategory.
Table~\ref{table:retrieval_res} shows precision-recall results for similar products retrieval. For evaluation, we considered furniture subcategories (42 in total) with number of images $\geq 20$.
\begin{table}[t]
%\vspace{-15pt}
\setlength\tabcolsep{4.3pt}
\centering
\caption{Precision/Recall results for image retrieval}
\label{table:retrieval_res}
\begin{tabular}{ c|c|c|c } 
\toprule
 & @1 & @3 & @5 \\ \midrule \midrule
precision@k & 0.468 & 0.409 & 0.374 \\ 
recall@k & 0.468 & 0.664 & 0.734 \\
\bottomrule
\end{tabular}
\vspace{-8pt}
\end{table}

\paragraph*{Generation quality:} we measure the performance of our copy-paste staging results by evaluating Frechet inception distance (FID) \cite{heusel2017gans}. FID is a popular metric for evaluating the quality of images created by GANs. The Wasserstein-2 distance in FID is calculated by comparing the features distribution of in-painted images with the distribution of real images, where the features are generated by a pre-trained InceptionV3 model. The comparison results are shown in Table~\ref{table:FID}. Since the copy-paste method in-paints only small regions of image around an object, it achieves much better FID score than vanilla staging. WBL further improves FID score in both methods.

\begin{table}[t]
%\vspace{-15pt}
\setlength\tabcolsep{4.3pt}
\centering
\caption{FID scores for various staging and training options}
\label{table:FID}
\begin{tabular}{c|c|c}
\toprule
 & EdgeConnect         & EdgeConnect \\ 
&  (baseline)          &  + WBL \\ \midrule \midrule
vanilla staging & 127.77 & 122.22  \\ %\midrule
copy-paste staging & 38.44 & 37.44 \\
\bottomrule
\end{tabular}
\vspace{-8pt}
\end{table}

\subsection{Human evaluation}
For a human perceptual study, we performed pairwise comparison tests, where experts were given two images at once, and the task was to determine which image appears more realistic (natural). For each comparison, we used three independent expert judgements, and decided the winner by majority voting. We performed three such tests: 1) 100 comparisons of vanilla staging images vs. ground truth images; 2) 100 copy-paste staging (with WBL) vs. ground truth; 3) 100 copy-paste (with WBL) vs. vanilla staging. The study showed:
\begin{itemize}
    \item 0\% of pix2pix images were better than ground truth;
    \item 3\% of copy-paste images were better than ground truth;
    \item 76\% of copy-paste images were better than pix2pix.
\end{itemize}
The above results clearly demonstrate the superiority of copy-paste staging and are in line with offline FID scores.

%\subsubsection{Editorial review}
%\subsection{Online results}
\section{Discussion} \label{sec:discussion}
Our proposed approach provides low budget advertisers a way to stage products digitally without having to spend on the physical resources needed for staging. Staging a room can easily cost up to few hundred dollars for an advertiser, and with image generation methods like the ones we have proposed, this would be basically free of cost (except for the legalities around copying backgrounds from other images). Leveraging the recent progress in prompt based image generation models, our approach can be further improved along the following lines: backgrounds from similar images could be used to generate prompts which then generate the background of the original product image. In addition, staged ads and parallax animations are expected to drive user engagement, and validating such hypothesis via an A/B test is one of our next steps. 
{\small
\bibliographystyle{ACM-Reference-Format}
\bibliography{refs}
}



%\maketitle

\end{document}